\documentclass{article}

\usepackage{microtype}
\usepackage{graphicx}
\usepackage{subfigure}
\usepackage{booktabs} 
\usepackage{scalerel}
\usepackage[hyphens]{url}
\usepackage{hyperref}

\usepackage[accepted]{icml2024}

\usepackage{amsmath}
\usepackage{amssymb}
\usepackage{mathtools}
\usepackage{amsthm}
\usepackage{xcolor}
\newcommand{\scalefact}{\gamma}
\newcommand{\cali}[1]{\textcolor{black}{#1}}
\usepackage[capitalize,noabbrev]{cleveref}

\theoremstyle{plain}
\newtheorem{theorem}{Theorem}[section]
\newtheorem{proposition}[theorem]{Proposition}

\newtheorem{corollary}[theorem]{Corollary}
\theoremstyle{definition}
\newtheorem{definition}[theorem]{Definition}

\theoremstyle{remark}

\usepackage[textsize=tiny]{todonotes}

\icmltitlerunning{On the Weight Dynamics of Deep Normalized Networks}

\begin{document}
	
	\twocolumn[
	\icmltitle{On the Weight Dynamics of Deep Normalized Networks}
	
	
	
	\icmlsetsymbol{equal}{*}
	
	\begin{icmlauthorlist}
		\icmlauthor{Christian H.X. Ali Mehmeti-Göpel}{unimz}
		\icmlauthor{Michael Wand}{unimz}
	\end{icmlauthorlist}
	
	\icmlaffiliation{unimz}{Department of Computer Science, Johannes-Gutenberg University, Mainz, Germany}

	\icmlcorrespondingauthor{Christian H.X. Ali Mehmeti-Göpel}{chalimeh@uni-mainz.de}
	
	\icmlkeywords{Machine Learning, ICML}
	
	\vskip 0.3in
	]
	
	
	
	\printAffiliationsAndNotice{\icmlEqualContribution} 
	
\begin{abstract}
	Recent studies have shown that high disparities in effective learning rates (ELRs) across layers in deep neural networks can negatively affect trainability. We formalize how these disparities evolve over time by modeling weight dynamics (evolution of expected gradient and weight norms) of networks with normalization layers, predicting the evolution of layer-wise ELR ratios. We prove that when training with any constant learning rate, ELR ratios converge to 1, despite initial gradient explosion. We identify a ``critical learning rate" beyond which ELR disparities widen, which only depends on current ELRs. To validate our findings, we devise a hyper-parameter-free warm-up method that successfully minimizes ELR spread quickly in theory and practice. Our experiments link ELR spread with trainability, a relationship that is most evident in very deep networks with significant gradient magnitude excursions.
\end{abstract}

\section{Introduction}
In the past decade, combining neural networks and big data has enabled dramatic breakthroughs \citep{alexnet,gpt4}, and network \textit{depth} has been a key factor: Compositions of many individual layers provide rich function spaces that  empirically appear to be better-aligned with real-world data distributions than any other inductive biases we are aware of today.
A fundamental problem of deep networks, maybe easily brushed over as technicality at first sight, is the problem of vanishing and exploding gradients. Propagating signals through a multi-layer networks is not easy: In the forward pass, the magnitude input signals easily increases or decreases, thus leading to an exponential excursion of \textit{signal magnitude}. Similarly, during the backward pass, we easily obtain similar excursion of \textit{gradient magnitudes}  \cite{meanfieldbn}. Further, using deep stacks of layers also easily increase correlations, thereby causing \textit{vanishing dimensionality} \cite{saxe2013exact}. Proper Initialization \cite{HeInit2015} can reduce the problem; trying to prevent it completely in a simple feed-forward network is challenging though \cite{Pennington-isometry-NIPS2017}.

Modern architectures \cite{resnet,transformer} thus usually address these issues by combining \textit{residual connections} \cite{resnet} and variants of \textit{normalization layers} such as \textit{batch normalization (BN)} \cite{batchnorm}. \cali{The former implicitly performs a down-weighting of deep paths, exponentially with depth \cite{VeitResNet2016}, and in combination with normalization layers, this effect is further increased (at initialization) by decreasing the weight of the residual branch \cite{De2020-BNRES}.}
The central objective of our paper is to understand how the \textit{dynamics} (over training time) of gradient \textit{magnitude} excursions (we do not consider correlations) are affected by normalization layers (BN and the similar).

\section{Related Work and Contributions}
Understanding of the benefits of BatchNorm \textit{standalone} is not straightforward and still subject to debate. The initial claim of reduced ``internal covariate shift'' was quickly refuted \citep{revisiting_covshift} and many alternative explanations were proposed, such as smoothing of the loss surface \citep{bn_smoothing} or enabling bigger learning rates \citep{bn_bigsteps}. 
Salimans and Kingma \yrcite{weightnorm} introduced WeightNorm, a method to decouple a layer's length and direction by training them as independent network parameters. They also demonstrated that in weight-normalized networks, gradients are orthogonal to layer weights, allowing update size calculation via the Pythagorean theorem. Hoffer et al. \yrcite{norm_matters} showed that the ``effective step size" in normalized networks is approximately proportional to $\frac{1}{||W||_2^2}$; this shows that scale invariance gives us an additional degree of freedom, as scaling a layer's weights is equivalent to inversely scaling its gradients or learning rate. You et al. \yrcite{lars} have observed that the ratio $\frac{||\nabla W||_F}{||W||_F}$, which we call effective learning rates (ELR), can vary wildly (up to a factor $\sim 250$ in AlexNet-BN) between layers after only one step of gradient descent. The authors conjecture that this can create instability in training, especially for large batch training requiring high learning rates, and propose to re-scale gradients by their effective learning rate. You et al. \yrcite{lamb} have later proposed a modified version of this algorithm for increased performance with transformer models. Brock et al. \yrcite{agc} have combined a similar re-scaling of the gradients with gradient clipping and are able to train normalizer-free networks using this technique. Bernstein et al. \yrcite{fromage} \cali{supported this intuition by showing that in a perturbed gradient descent, an optimization step decreases the loss function if all layer-wise ELRs are bounded by a term that depends on the perturbation angle.}  Arora et al. \yrcite{autotune} described the auto rate-tuning effect, proving \cali{that gradient descent} asymptotically converges to a stationary point without manual tuning of learning rates for specific layers, given certain assumptions. \cali{Wan et al. \yrcite{smd} prove that the ``angular update" (a measure similar to ELR) of a given normalized layer eventually converges to a constant limit value which does not depend on initial conditions, but rely on weight decay for their demonstration. Interestingly, Li and Arora \yrcite{exp_lr_wd} show that using weight decay with a constant learning rate schedule is mathematically equivalent to using no weight decay and an exponentially increasing learning rate schedule.}

\cali{The above-mentioned works show that the learning speeds of different layers do eventually align, but glance over the importance of correct learning rate scheduling in the early training phase, which we believe to be crucial in practice. We find that while convergence is always guaranteed given simplifying assumptions and over an indefinite number of iterations, choosing an excessively high learning rate, especially in the first steps of training, can drastically increase imbalances in layer-wise learning speeds (ELR spread) to a degree where recovery is impossible within a realistic time frame. Furthermore, Li and Arora show a connection between weight decay and warm-up but do not demonstrate how these techniques affect ELR spread. }

In this work, we model the dynamic effects solely induced by normalization layers and assume that the layer-wise gradient magnitude excluding normalization effects (base gradient magnitude) remains constant over time. In this setting, we derive a model predicting the evolution of a network's weight dynamics (expected layer-wise gradient and weight norms). In the gradient flow, this behavior reduces to a non-linear ODE with a closed-form solution, where all ELR ratios \cali{between layers}  smoothly converge to 1. When training with higher learning rates, the behavior changes fundamentally, as the layer with the highest ELR can flip even below the layer with the lowest ELR in a single step if a certain \textit{critical learning rate} is exceeded, which in turn increases ELR spread of the network. When training with constant learning rates, ELR \cali{spread}  can increase only during the first step, slowing down convergence, but still eventually converging. From there, we derive a warm-up scheme that is guaranteed to converge in $num\_layers$ steps. Empirically, we were able to show that high ELR spreads indeed seem to \cali{correlate with low} trainability: by using techniques that control ELR spread (gradient normalization and warm-up), we are able to reduce the high (initially exponential in the number of layers) ELR spreads of a 110 layer feedforward network and render the previously un-trainable network trainable. In summary, we create a theoretical framework that shows how the dynamical effects of normalization layers can help counter gradient magnitude excursions in deep neural networks.

\section{Auto Rate-Tuning Effect and Its Dynamics}
The core observation is that for any layer $N$ that is invariant to scaling in the forward pass $N(\scalefact \cdot x) = N(x)$ (e.g. all normalization layers), its gradient scales inversely with its input:
\begin{equation}
	\frac{dN}{d \scalefact x}(\scalefact x) = \frac{1}{\scalefact}\frac{dN}{dx}(x).
	\label{inv_scaling}
\end{equation}
This is a simple consequence of the chain rule and has been shown for BatchNorm by Wu et al. \yrcite{wngrad} and for LayerNorm by Xiong et al. \yrcite{onlayernorm}. 
Secondly, Arora et al. \yrcite{autotune} show that since normalization layers are scale-invariant, no gradient can flow in this direction. Hence, weight updates $\nabla W$ are orthogonal to the weights $W$ themselves:
\begin{equation}
	\langle\nabla W,W\rangle = 0.
	\label{orthogonality}
\end{equation}
We now explore how this affects a network's weight dynamics. Intuitively, the weight norm of layers with high gradient norms grows fast and thus down-scales the gradient, leading to auto-regulation: this effect is called \textit{auto rate-tuning}. We would like to point out that in a realistic scenario, the data tensor is multi-dimensional and condition \ref{inv_scaling} is satisfied along a subset of its dimensions (e.g. the batch, height and width axis for BatchNorm); auto-rate tuning is therefore given along those dimensions.

\subsection{Sufficient Conditions for Auto Rate-Tuning / Correct Placement of Normalization Layers}
\label{suff_cond_autotune}
A necessary condition for auto rate-tuning of a linear layer $L$ is the invariance of the network's output with respect to re-scaling the weights in $L$  \citep{autotune}. We deduct that any type of normalization layer (e.g. BatchNorm, LayerNorm) induces auto-rate tuning and that placing a normalization layer directly after every linear layer, as it is the case in most convolutional networks, is sufficient to achieve scale-invariance. In Transformer models, this was initially not the case, and we conjecture that this could explain the improvements when adding additional normalization layers in the feedforward blocks \citep{normformer} or query/key blocks \citep{qk_norm}. Arora et al. also note that the scale invariance property is not disrupted by positive homogeneous functions of degree 1. We infer the following classification of commonly used layers:
\begin{center}
	\begin{tabular}{ l l }
		Auto-tuning passes through & Breaks auto-tuning \\
		\toprule 
		Linear layers w/o bias & Linear layers w. bias\\
		Homogeneous nonlin. of deg. 1 &Other nonlinearities\\
		Dropout & MaxPool \\
	\end{tabular}
\end{center}
If a residual connection is placed in-between a linear layer and the next normalization layer, it can break auto rate-tuning; this is the case e.g. in a ResNet v2 \citep{resnetv2}. 

\subsection{Training Dynamics Induced by Auto Rate-Tuning}
To model training dynamics, we assume that weights of a given layer, as well as their gradients, are random matrices where entries are normally distributed with zero mean and a time-dependent standard deviation that is uniform in each layer.
%
%
We parameterize training time $t \in \mathbb{R}$ such that $t_i=i\cdot\lambda^2$ after $i$ optimization steps with a constant learning rate $\lambda > 0$. In this notation, gradient descent updates can be written as $W(t_{i+1})=W(t_i+\lambda^2) = W(t_i) - \lambda \nabla W(t_i)$. The updates preserve zero norm and uniform variance of all entries in $W$. Assuming the independence of all entries in the weight and gradient matrices, we can deduce the following update rule from the orthogonality condition (\ref{orthogonality}): 
\begin{equation}
	||W(t_{i+1})||^2_F = ||W(t_i)||_F^2 + \lambda^2||\nabla W(t_i)||_F^2.
\end{equation}
Condition (\ref{inv_scaling}) implies that gradient updates are inversely proportional to the current layer weights. We now assume that the ``base gradient" of a layer, meaning the gradient magnitude excluding normalization induced scaling effects, is constant during training i.e. 
\begin{equation}
	\label{eq:ctt_base_grads}
	\mathbb{E}(||W(t_i)||_F \cdot ||\nabla W(t_i)||)_F=c,
\end{equation}
for a constant $c\in\mathbb{R}$ at all times-steps $t_i$. We discuss the limitations of this assumption in Section \ref{sec:limitations}. Using shorthand $\sigma^2(t_i)\coloneq\mathbb{E}(\|W(t_i)\|^2_F)$ and $\sigma(t_i) =\sqrt{\sigma^2(t_i)}$, we obtain:
\begin{equation}
	\sigma^2(t_{i+1}) = \sigma^2(t_i) + \frac{\lambda^2 c^2}{\sigma^2(t_i)}, \label{discrete_model_weights}
\end{equation}
for a constant base gradient $c > 0$ depending only on layer depth and initial weights norm that we assume to be strictly positive $\sigma^2(0) > 0$. We call this the \textbf{discrete model}.

\subsection{Gradient Norms at Initialization}
\label{expl_grads}
\textbf{Feed-forward networks:} The dynamics of Eq.~\ref{weight_norm_eq} apply to all normalized layers equally, but the initial gradient norms $\|\nabla{W}_i(0)\|_F$ differ substantially across layers $i \leq L$: Yang et al. \yrcite{meanfieldbn} show that in feedforward networks with Batch Normalization, the gradient norm at initialization grows as:
\begin{equation}
	c_i \sim \alpha^{L - i},
	\label{expl_gradnorm}
\end{equation}
with $\alpha \coloneq \sqrt{\pi/(\pi-1)} \approx 1.21$ for ReLU activations and He. initialization. See also Luther \yrcite{luther2020} for a simplified derivation.

\textbf{ResNets:} When considering residual networks, as per the multivariate chain rule, the gradient of residual blocks is additive instead of multiplicative \citep{resnetv2}. Additionally, frequency-dependent signal-averaging further dampens gradients in a ResNet \citep{ringing_relus}. It follows from the consideration for fully-connected network above and He et al. \yrcite{resnetv2} Eq. 5 that for a residual network using ReLU units:
\begin{equation}
	c_i \sim 1 + \lfloor \frac{L-i}{s}\rfloor\alpha^s,
	\label{expl_gradnorm_resnet}
\end{equation}
where $s$ is the number of ReLU units in a residual block.
\subsection{Auto Rate-Tuning Affects Each Layer Separately}
In this section, we establish that the dynamic re-scaling of gradients explored above applies to each layer independently and does not affect layers above or below, showing that a simple layer-wise view is sufficient.

\begin{proposition}[Every Layer Auto-Tunes Separately]
	\label{prop:independent_layers}
	Consider a concatenation of a linear layer $L(x, W) = x^T W$  followed by a normalization layer $N$. Then, the derivative wrt. the input remains the same when layer weights are scaled by a factor $\gamma$:
	\begin{equation}
		\frac{dN}{dx} (x, \scalefact W) = \frac{dN}{dx} (x, W).
	\end{equation}
\end{proposition}
The proof can be found in the Appendix Section \ref{appendix:sec:additional_proofs}.

\subsection{Effective Learning Rates and Their Ratios}
\cali{To account for scale variance induced by normalization layers, we are interested in the update size of the weight direction  $\widehat{W} \coloneq \frac{W}{~\left\lVert W \right\rVert_2}$. Similarly to van Laarhoven \yrcite{l2_vs_wn}, by approximating $\lVert W(t_{i+1})\rVert_2 \approx \lVert W(t_{i})\rVert_2$, we can write:
\begin{equation}
	\widehat{W}(t_{i+1})-\widehat{W}(t_{i}) \approx \frac{W(t_{i+1}) - W(t_{i}) }{\lVert W(t_{i})\rVert_2} \sim \frac{\nabla W(t_i)}{\lVert W(t_{i})\rVert_2} .
\end{equation}}
It is therefore imperative to consider the ratio from gradient-to-weight norm as measure of change in the layer's weights.
\begin{definition}[Effective Learning Rate]
	We define the effective learning rate $E$ of a layer with weight norm $\sigma^2$ and base gradient $c$ as:
	\begin{equation}
		E(t_i) \coloneq \mathbb{E}\left(\frac{||\nabla W(t_i)||_F}{||W(t_i)||_F}\right)=\frac{c}{\sigma^2(t_i)}.
	\end{equation}
\end{definition} 
As all effective learning rates can simply be globally re-scaled by adjusting the learning rate, we are interested in the evolution of layer-wise ratios of effective learning rates.
\begin{definition}[Effective Learning Rate Ratios]
	We define the effective learning rate ratio $R_{j k}$ of two layers $j$ and $k$ with weight norms $\sigma_j^2, \sigma_k^2$ and base gradients $c_j, c_k$ at a given time step $t_i $ as:
	\begin{equation}
		R_{j k}(t_i) \coloneq \frac{E_j}{E_k} (t_i) = \frac{c_j \sigma_k^2}{c_k\sigma_j^2}(t_i) .
		\label{def:ratio}
	\end{equation}
\end{definition}

\subsection{Analysis in the Gradient Flow}
In this section, we show that in the gradient flow, weight dynamics have a closed-form solution and all ELR ratios converge smoothly to 1. 
\begin{theorem}[Closed-Form Solution]
	In the gradient flow ($\lambda\rightarrow0$), Eq. \ref{discrete_model_weights} has the following closed form solution:
		\begin{equation}
		\sigma^2(t) = \sqrt{2 c^2 t+k_0}.\\
		\label{weight_norm_eq}
	\end{equation}
 with $k_0 = 4$, assuming He initialization \citep{he_init}. We will further call this the \textbf{continuous model}. 
\end{theorem}
\begin{proof}
	Starting from Eq. \ref{discrete_model_weights}, we can utilize that $t_{i+1}=t+\lambda^2$ to drop the index and solve for the difference quotient:
	\begin{equation}
		\frac{\sigma^2(t+\lambda^2) - \sigma^2(t)}{\lambda^2} = \frac{c^2}{\sigma(t)^2}.
	\end{equation}
	In the limit $\lambda^2 \rightarrow 0$, this yields the gradient flow that can be expressed as a nonlinear first order differential equation :
	\begin{equation}
		\frac{d \sigma^2}{dt} = \frac{c^2}{\sigma^2}
		\label{def_grad}
	\end{equation}
	The exact positive solution to the differential equation is given by:
	\begin{equation}
		\sigma^2(t) = \sqrt{2 c^2 t+k_0}.\\
	\end{equation}
	Assuming He initialization, the expected initial squared weight norm is $2$ for layer width $n$. Thus, $2 =\sigma^2(0) = \sqrt{k_0}$ and therefore $k_0 = 4$.
\end{proof}

\begin{theorem}[Convergence to Fixed Point]
	 In the gradient flow ($\lambda\rightarrow0$), all effective learning rate ratios eventually converge given enough time, i.e. for any layer pair $j,k\leq L$:
	\begin{equation}
		\lim_{i\rightarrow \infty}R_{j k}(t_i)=1.
	\end{equation}
\end{theorem}

\begin{proof}
	We consider two arbitrary layers $j$ and $k$ with respective weight norms  $\sigma_j^2$, $\sigma_k ^2> 0$ and base gradients $c_j$, $c_k > 0$. Using the formulae for gradient norm (Eq. \ref{def_grad}) and weight norm (Eq. \ref{weight_norm_eq}) in the continuous model, we write:	
	\begin{equation}
		\frac{E_j}{E_k}(t) = \frac{c_j}{\sigma^2_j(t)} \cdot \frac{\sigma^2_k(t)}{c_k}
		= \frac{c_j \sqrt{2 c_k^2 t+ k_0}}{c_k\sqrt{2 c_j^2 t+ k_0}}
		\overset{t\rightarrow\infty}\longrightarrow 1.
		\label{eq:ratios}
	\end{equation}
\end{proof}

%
\subsection{Analysis for Bigger Learning Rates}

\label{sec:theory_largelr}
In this section, we characterize the evolution of ELR ratios for non-infinitesimal, scheduled learning rates $\lambda(t_i)$, now relying solely on the discrete model. If $\lambda(t_i)$ is constant, we find the asymptotic behavior to be the same as in the gradient flow, where ELRs ratios converge in the time limit. On the contrary to the continous model, ratios can (temporarily) widen when surpassing a certain critical learning rate.

\begin{theorem}[Convergence to Fixed Point]
	\label{theorem:convergence_bigger_lr}
	In the time limit and for a constant learning rate $\lambda(t_i)=\lambda$, all effective learning rate ratios converge. For any layer pair $j,k\leq L$:
	\begin{equation}
		\lim_{i\rightarrow \infty}R_{j k}(t_i)=1.
	\end{equation}

\end{theorem}
The proof can be found in Appendix Section \ref{appendix:sec:additional_proofs}. The main idea is  that by substituting $x_i\coloneq\frac{\sigma^2_j(t_i)}{c_j\lambda}$ and $y_i\coloneq\frac{\sigma^2_k(t_i)}{c_k\lambda}$, we can rewrite Eq. 4 for two distinct layers $j$ and $k$ as two sequences obeying the same recurrence relation and consequently bound the expression.

\begin{proposition}[Ratios Shrink]
	Let $j, k \leq L$ be any layer pair.
	\begin{enumerate}
		\item 	If $R_{j k} (t_i)> 1$, the ratio $R_{j k}$ is then strictly lower in the next time step, i.e.	$R_{j k}(t_{i+1}) < R_{j k}(t_i).$
		\item 	If $R_{j k} (t_i)< 1$, the ratio $R_{j k}$ is then strictly greater in the next time step, i.e.	$R_{j k}(t_{i+1}) > R_{j k}(t_i).$
	\end{enumerate}
	\label{lemma:statpt}
\end{proposition}

\begin{proof}
	We start by showing the first proposition. We can reformulate the expression $\frac{E_j}{E_k}(t_{i+1}) < \frac{E_j}{E_k}(t_{i})$  using the definition of the effective learning rate as the following equivalent expression:
	\begin{equation}
		\frac{c_j^2 \sigma_k^4}{\sigma_j^4 c_k^2} (t_{i+1}) <	\frac{c_j^2 \sigma_k^4}{\sigma_j^4 c_k^2} (t_{i}).
	\end{equation}
	We simplify this expression and take the square root. Since $\sigma$ are variances and thus non-negative, the following expression is also equivalent:
	\begin{equation}
		\frac{\sigma^2_k} {\sigma^2_j}(t_{i}) - \frac{\sigma^2_k}{\sigma^2_j} (t_{i+1}) > 0.
		\label{lemma1:toshow}
	\end{equation}
	We expand the second term using Eq. \ref{discrete_model_weights}: 
	\begin{align}
		\frac{\sigma^2_k}{\sigma^2_j} (t_{i+1})  &= 	\frac{\sigma^2_k + \frac{c_k^2\lambda^2}{\sigma^2_k}}{\sigma^2_j + \frac{c_j^2\lambda^2}{\sigma^2_j}}(t_i)
		= \frac{\sigma^2_j\sigma^4_k +\sigma^2_j c_k^2\lambda^2}{\sigma^4_j\sigma^2_k + \sigma^2_kc_j^2\lambda^2} (t_i).
		\label{eq:nextstep}
	\end{align}
	
	Substituting this term on the left hand side of Eq. \ref{lemma1:toshow} and combining the terms yields:
	\begin{align}
		\frac{\sigma^2_k}{\sigma^2_j}(t_{i})  -  \frac{\sigma^2_j\sigma^4_k +\sigma^2_j c_k^2\lambda^2}{\sigma^4_j\sigma^2_k + \sigma^2_kc_j^2\lambda^2} (t_i) 
		&= \frac{\sigma^4_k c_j^2\lambda^2 -\sigma^4_j c_k^2\lambda^2}{\sigma_j^2(\sigma^4_j\sigma^2_k + \sigma^2_kc_j^2\lambda^2)} (t_i).
	\end{align}
	Since the denominator is strictly positive. Eq. \ref{lemma1:toshow} is therefore equivalent to:
	\begin{equation}
		\sigma^4_k c_j^2\lambda^2 (t_i) > \sigma^4_j c_k^2\lambda^2  (t_i),
		\label{eq:sp_condition}
	\end{equation}
    which is in turn equivalent to $\frac{E_j }{E_k} (t_{i}) > 1$   by definition. The second proposition can be shown analogously.
\end{proof}

A consequence of this proposition is that a given ratio $R_{jk}$ diminishes during every step, except for when it flips, i.e. $R_{jk}(t_i) > 1$ and $R_{jk}(t_{i+1}) < 1$. In Appendix Section \ref{appendix:sec:additional_proofs}, we show that when training with constant learning rates, this can only happen during the first step. Now, we would like to find the precise learning rate where the ratio flips.

\begin{definition}[Flipping Ratio]
	We define the ``flipping ratio" $\kappa_{jk}$ of two layers $j$ and $k$ at a given time step $t_i$ as:
	
	\begin{equation}
		\kappa_{j k} (t_i)\coloneq \frac{\sigma_j\sigma_k}{\sqrt{c_j c_k}}(t_i)=\sqrt{\frac{1}{E_j E_k}}(t_i).
	\end{equation}
\end{definition}

\begin{proposition}[Flipping Conditions]
	\label{lemma:flipping_conditions}
	Let $j, k \leq L$ be any layer pair with w.l.o.g. be $R_{j k}(t_i)> 1$. 
	
	\begin{enumerate}
		\item The effective learning rate ratio does not flip between time steps $t_i$ and $t_{i+1}$ i.e. $ R_{j k}(t_{i+1})> 1$ if and only if $	\lambda (t_i)<  \kappa_{j k}(t_i)$.
		\item The effective learning rate ratio does flip between time steps $t_i$ and $t_{i+1}$ i.e. $ R_{j k}(t_{i+1})< 1$ if and only if $	\lambda (t_i) >  \kappa_{j k}(t_i)$.
		\item The ratio $\frac{E_j}{E_k}$ has reached a stationary point at a given time step $t_i$, i.e. $R_{j k}(t_{j}) = R_{j k}(t_{j+1})$ for all $j\geq i$ if and only if $\lambda (t_i) = \kappa_{j k}(t_i)$.
	\end{enumerate}
\end{proposition}

\begin{proof}
	We start by showing the first proposition. Using the definition of $R_{j k}$ and Eq. \ref{eq:nextstep}, we can write:
	\begin{align}
		R_{j k}(t_{i+1})=\frac{c_j \sigma^2_k}{c_k \sigma^2_j} (t_{i+1})  
		&=\frac{\sigma^2_j\sigma^4_k c_j +\sigma^2_j c_j c_k^2\lambda^2}{\sigma^4_j\sigma^2_k c_k + \sigma^2_k c_j^2 c_k \lambda^2} (t_i).
		\label{eq:ratio}
	\end{align}
	Thus, the condition $ R_{j k}(t_{i+1})> 1$ is equivalent to:
	\begin{align}
		&\left(\sigma^2_j\sigma^4_k c_j +\sigma^2_j c_j c_k^2\lambda^2\right)(t_i)
		>
		\left(\sigma^4_j\sigma^2_k c_k + \sigma^2_k c_j^2 c_k \lambda^2\right)(t_i) \\
		\Leftrightarrow&
		\left(\sigma^2_j c_j c_k^2\lambda^2  - \sigma^2_k c_j^2 c_k \lambda^2 \right)(t_i)
		>
		\left(\sigma^4_j\sigma^2_k c_k - \sigma^2_j\sigma^4_k c_j\right)(t_i)\\
		\Leftrightarrow&
		\lambda^2(\sigma^2_j c_j c_k^2  - \sigma^2_k c_j^2 c_k)  (t_i)
		>
		\sigma^4_j\sigma^2_k c_k - \sigma^2_j\sigma^4_k c_j  (t_i)\\		
		\Leftrightarrow&
		c_j c_k\lambda^2(\sigma^2_j c_k  - \sigma^2_k c_j)  (t_i)
		>
		\sigma^2_j\sigma^2_k (\sigma^2_j c_k - \sigma^2_k c_j)  (t_i)
	\end{align}
	Since we assumed  $ R_{j k}(t_{i}) = \frac{E_j}{E_k} (t_i)= \frac{\sigma^2_k c_j}{\sigma^2_j c_k}(t_i)> 1$, it follows that  $\left(\sigma^2_j c_k - \sigma^2_k c_j\right)(t_i) < 0$  and thus we invert the sign of the inequality when dividing by this quantity and we obtain the following equivalent condition:
	\begin{equation}
		\lambda^2  (t_i)
		<
		\frac{\sigma^2_j\sigma^2_k }{c_j c_k}(t_i). \\
	\end{equation}
	All quantities are non-negative, therefore taking the square root preserves equivalence and we obtain the sought condition. The other propositions can be shown analogously.
\end{proof}

Since we are interested in reducing the highest overall ratio $R_{h \ell}(t_i)$ where $\ell, h$ are the layers with the lowest respective highest effective learning rate, we call $\kappa_{\ell h}(t_i)$ the \textit{critrical learning rate}. When using higher learning rates than this value, $E_h(t_i)$ flips below $E_\ell(t_i)$ during the next step, thus (for high $\lambda$ considerably) increasing total ELR spread. In the following we will come to understand that in practice, a more conservative choice is advisable; for this reason, we propose the \textit{subcritical}, but still provably fast warm-up scheme below.

\begin{corollary}[Subcritical Warm-Up]
	\label{cor:supercritical}
	Given a network with $L>0$ layers, if we schedule the learning rate as $\lambda(t_i) = \kappa_{h h'}(t_i)$, where we chose $h, h' \leq L$ at each step to be the two layers with the highest effective learning rates, then no ratio $R_{j k}$ for any $j,k\leq L$ ever flips between a time step and the next 
	and all pairs of effective learning rate ratios $R_{j k}$ for any $j,k\leq L$ converge to 1 in $L$  steps.
\end{corollary}
\begin{proof}
	Let $h, h'$ be the two layers with the highest effective learning rates at time step $t_i$. By Proposition \ref{lemma:flipping_conditions}, if we chose  $\lambda(t_i) = \kappa_{h h'}(t_i)$, we have $R_{h h'}=1$ at the next time step $t_{i+1}$ and for all further time steps. Since  $h, h'$ are the two layers with the highest effective learning rate at time step $t_i$ and we chose $\lambda(t_i)$ to be equal to their flipping ratio $\kappa_{h h'}(t_i)$, we have:
	\begin{equation}
		\lambda(t_i) = \kappa_{h h'}(t_i) \leq  \kappa_{j k}(t_i)
	\end{equation}
	for all other layers $j,k\leq L$.  Therefore, by Lemma \ref{lemma:flipping_conditions}, no ratio $R_{j k}$ will ever flip between a time step and the next for any $j,k\leq L$. If we repeat this process $L$ times, all pairwise learning rate ratios converge to 1.
\end{proof}

\subsection{Simulating Warm-Up Schedulers and Criticality}

\begin{figure*}[t!]
	\centering
	\includegraphics[width=\linewidth]{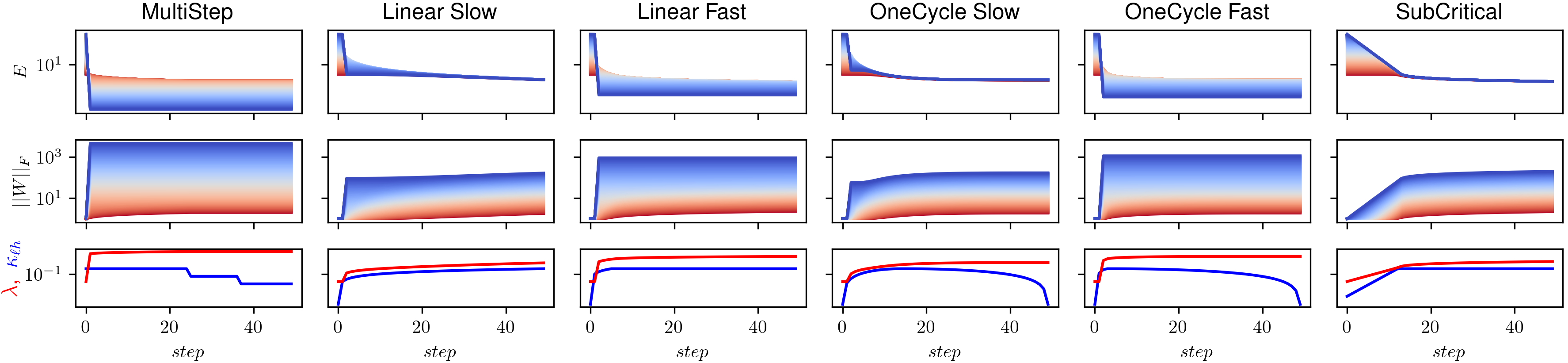}
	\caption{Simulated evolution of layer-wise effective learning rates \textbf{(top)}, weight norms  \textbf{(middle)}, learning rates $\lambda(t)$ and critical LRs  $\kappa_{\ell h}(t)$ \textbf{(bottom)} for popular learning rate schedulers. All y-axes are in logarithmic scale. Blue color corresponds to the lower layers.}
	\label{sched_compare}
\end{figure*}

In Figure \ref{sched_compare}, in order to visualize the concept of criticality, we simulated the evolution of effective learning rates and weight norms for popular learning rate schedulers with our discrete model (ref. Eq. \ref{discrete_model_weights}), assuming initially exponentially exploding gradients (ref. Eq. \ref{expl_gradnorm}). We also indicated $\lambda(t)$ along with the critical learning rate $\kappa_{\ell h}(t)$.  As predicted by our analysis in Section \ref{sec:theory_largelr}, whenever $\lambda(t)>\kappa_{\ell h}(t)$, we see that the highest effective learning rate flips below the lowest, which in turn increases the ELR spread and consequently the convergence time. We conclude that whether a warm-up scheme succeeds in quickly reducing ELR spread highly depends on the chosen hyper-parameters.

\section{Experimental Validation}
\begin{figure}[tbp]
	\centering
	\includegraphics[width=\columnwidth]{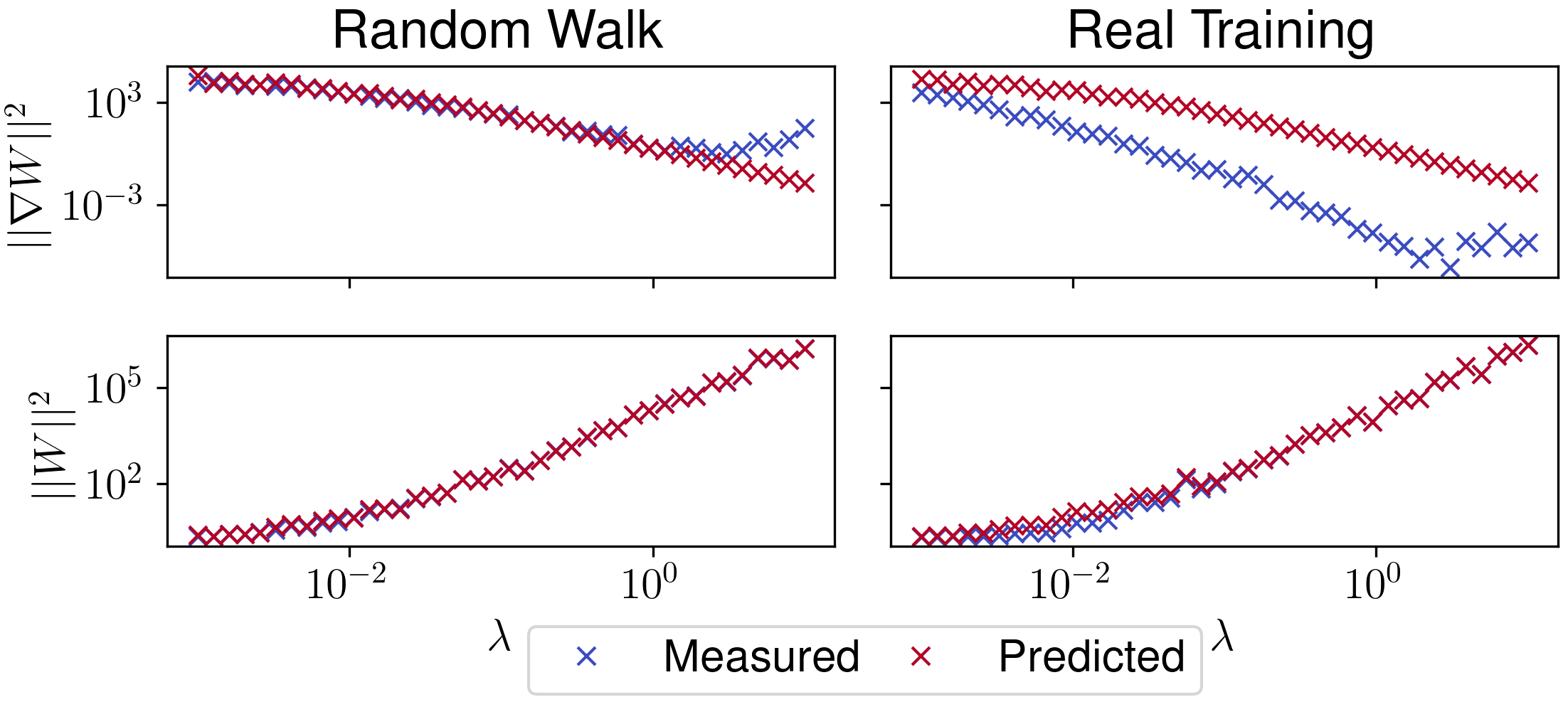}
	\caption{Short term evolution (after 10 steps) of predicted vs. measured  weight and gradient norms of the lowest layer of a ResNet56 NoShort trained with random gradients \textbf{(left)} and real gradients \textbf{(right)} for various $\lambda$.}
	\label{fig:sanity_shortterm}
\end{figure}

In this experimental Section, we will first check the limitations of the assumption about constant base gradients and validate the predictivity of our model. Then, we will compare the predicted critical learning rate to an empirical value extracted from real training runs. Finally, we confirm that high ELR spreads \cali{correlate} with network trainability in practice. 
\subsection{Experimental Setup}
\subsubsection{Architectures, Datasets and Training Protocols}
 We chose ResNet v1 \citep{resnet} with (``Short") and without (``NoShort") residual connections as examples of standard architectures. We chose a ResNet v1 as opposed to a v2 since in the former, the ``correct'' placement of normalization layers (ref. Section \ref{suff_cond_autotune}) is given without modifying the architecture. Theory predicts that a high number of layers and not using residual connections increase the strength of the observed effect (ref. Section \ref{expl_grads}). We therefore use 56 and 110 layer networks: Without residual connections, the former is deep but still trainable and the latter is mostly un-trainable with basic constant LR training. \cali{The final layer of a ResNet v1 is not scale-invariant and we therefore exclude it from our analysis.}
 For computer vision tasks, we work with standard image classification datasets of variable difficulty: CIFAR-10, CIFAR-100  \citep{cifar} and ILSVRC 2012 (called ImageNet in the following) \citep{imagenet}. We use the most basic training setting possible (vanilla SGD) and disable all possible factors that influence weight dynamics: momentum, weight decay, affine BatchNorm parameters and bias on linear layers (for a discussion, please refer to Appendix Section \ref{appendix:sec:limitations}). We further use different kinds of learning rate scheduling with and without warm-up; further details about the architectures and training process can be found in the Appendix.

\subsubsection{Measuring ELR Spread}
In our experiments, we need a measure for ELR spreads that is relative to the network's mean ELR.
\begin{definition}[Relative Logarithmic ELR Spread]
	We define the Relative Logarithmic ELR Spread as:
	\begin{equation}
		S_{rel}\coloneq\text{std}(\text{ln}(E)),
	\end{equation}
	computed across the layers of the network and usually averaged over all channels and the entire training process.
\end{definition}

\begin{figure*}[tbp]
	\centering
	\includegraphics[width=\linewidth]{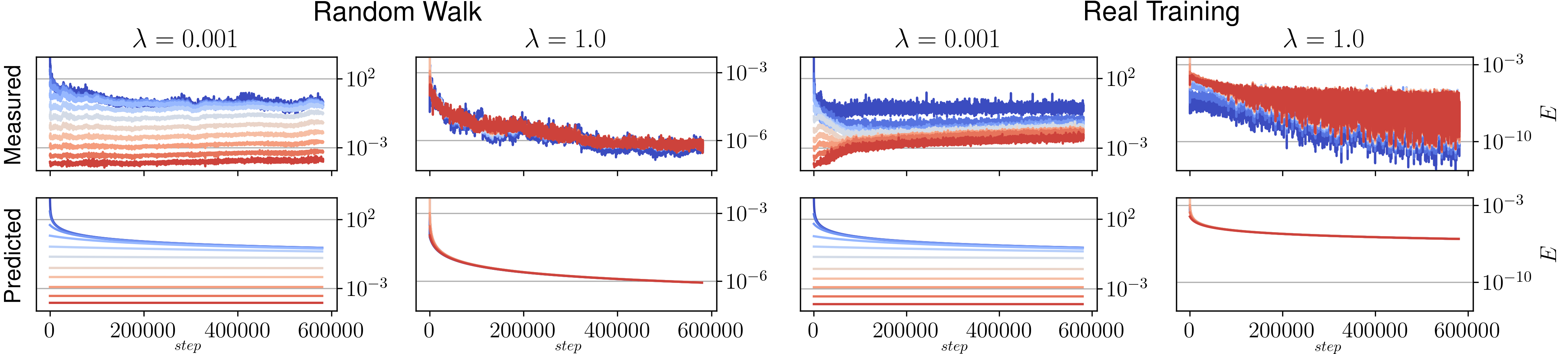}
	\caption{Long term evolution of predicted  vs. measured layer-wise effective learning rates for a ResNet56 NoShort trained with random gradients \textbf{(left)} and real gradients \textbf{(right)}. Blue color corresponds to the lower layers.}
	\label{fig:pred_drift}
\end{figure*}
\label{sec:limitations}

\subsubsection{Random Walk}
In the past section, we modeled exclusively the dynamics induced by normalization assuming constant base gradients (ref. Eq \ref{eq:ctt_base_grads}), meaning that the layer-wise expected gradient magnitude excluding normalization effects is constant over time. This is obviously not strictly true in a practical setting: apart from the obvious factors mentioned above (momentum, affine parameters etc.), the derivative of nonlinear layers and the objective function itself changes when varying inputs or weights. During training, gradient norms tend to shrink as the objective function saturates \citep{wide_nets}. \cali{To verify that mostly learning effects are responsible for fluctuations in base gradients, we observe how weight dynamics evolve during a random walk.}

\begin{definition}[Random Walk]
	\cali{During each training step, before applying the gradients computed in the backward-pass, we replace every layer's gradient by a random vector of similar norm which is also orthogonal to the layer's weights. Please refer to Algorithm \ref{alg:rand_walk} for a formal description.}
	\begin{algorithm}[htb]
		Let $e_\ell$ denote the number of elements of the weight vector $W_\ell$ and $\langle \cdot,\cdot \rangle$ the dot product. 
		\caption{Random Walk}
		\label{alg:rand_walk}
		\begin{algorithmic}
			\FOR{each gradient descent step $i$}
			\FOR{each layer $\ell$}
			\STATE Compute $\nabla W_\ell(t_i)$
			\STATE $\sigma \leftarrow \sqrt{ \lVert \nabla W_\ell (t_i) \rVert_2^2 / e_\ell}$
			\STATE $R \leftarrow \mathcal{N}(0, \sigma^2)^{e_\ell}$ 
			\STATE $V \leftarrow W(t_i)$
			\STATE $\nabla W_\ell(t_i) \leftarrow R - \frac{\langle R,V\rangle}{\langle V,V \rangle} V$ // orthogonalize
			\ENDFOR
			\ENDFOR
		\end{algorithmic}
	\end{algorithm}
\end{definition}

\subsection{Model Validation and Limitations}

To validate our \cali{theory, we measure the initial gradient and weight norms of a network and extrapolate their evolution using our discrete model (Eq. \ref{discrete_model_weights}). We then compare the predicted} weight/gradient norms to the \cali{empirically measured} values after a given number of steps. We will first see that for a feedforward network with ReLU activations, it is already enough to exclude learning effects (\cali{random walk scenario}) for our model to be long-term predictive. When including them, as expected, gradients are lower than predicted but the main takeaway qualitatively still holds: \cali{ELR spreads diminish over time, given that a certain learning rate is not exceeded}. 

 \textbf{Short-Term Validation :} In Figure \ref{fig:sanity_shortterm}, we compare the \cali{measured weight/gradient norm of the lowest layer of a ResNet56 NoShort after training on Cifar10 for 10 steps to the values predicted using our discrete model on the initial values. In a random walk} (left), predictions are quite accurate up to $\lambda\approx1$ and get slightly inaccurate for higher $\lambda$, presumably due to numerical issues. \cali{As for real training} (right), we see that gradients are notably smaller than expected after 10 steps. For the following, it is crucial to note that the difference between the predicted and \cali{measured} values is not a constant ratio but instead increases in $\lambda$.

\begin{figure*}[tbp]
	\centering
	\includegraphics[width=\linewidth]{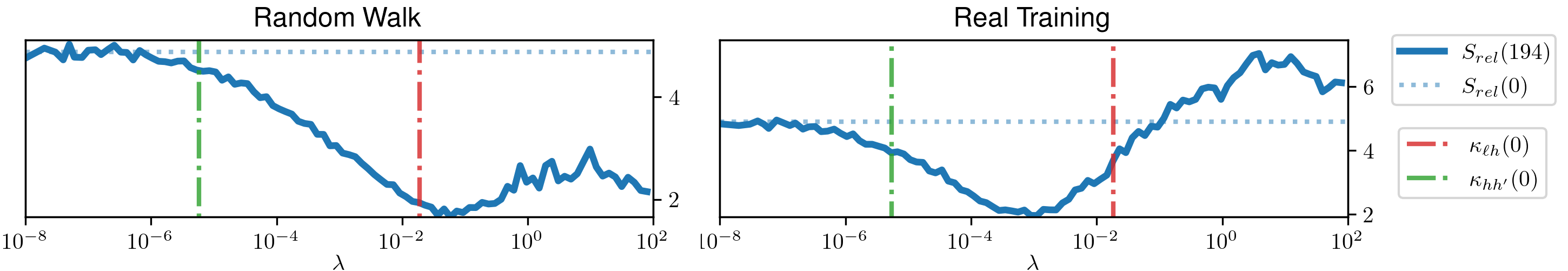}
	\caption{Relative spread after one epoch \textbf{(solid blue)}, relative spread at initialization \textbf{(dotted blue)} and the critical \textbf{(red)} / subcritical \textbf{(green)} learning rate at initialization of a ResNet110 NoShort with random gradients (\textbf{left}) and real gradients (\textbf{right}).}
	\label{fig:critical_lr}
\end{figure*}

\textbf{Long Term Validation :} We conducted a similar experiment for only two different learning rates $\lambda\in\{0.001,1\}$ over 3000 epochs and visualize the \cali{measured/predicted} ELR of all layers in Figure \ref{fig:pred_drift}. We see that \cali{in the random walk scenario}, our prediction is remarkably accurate. \cali{In real training} and for $\lambda=1$, our model predicts this learning rate to be critical, but in reality it is super-critical as the gradients of  the lower layers (blue) significantly undershoot with regard to the prediction and their ELR flips below the highest layers (red); further training does not seem to be able to recover the high ELR spread. \cali{Since training with any subcritical learning rate reduces ELR spread, we will see that it is sufficient to use a slightly lower $\lambda$ than the predicted critical value to avoid an increase in ELR spread.}

\textbf{Predicting Criticality:} In Figure \ref{fig:critical_lr}, we train a ResNet110 NoShort for a single epoch using various constant learning rates on Cifar10 \cali{in a random walk and a real training scenario, tracking the evolution of ELR spreads. We plot ELR spreads at initialization and after one epoch, averaging measurements over 10 runs for each datapoint. First, we note that as predicted, up to a certain learning rate, ELR spreads are always lowered by training. Next, we} indicate the predicted (sub)critical learning rate at initialization: as per Proposition \ref{prop:fliponce}, if a learning rate is subcritical in step 0, it should \cali{also }not increase spreads during later steps. Consequently, we expect the runs with $\lambda\approx\kappa_{\ell h}(0)$ to have the lowest $S_{rel}$ value \cali{after training}. In Figure \ref{fig:critical_lr}, we see that this is indeed the case \cali{for the random walk. In real training,} the qualitative behavior is similar but the curve is shifted to the left as gradients are smaller. We also note \cali{that in real training, when using} super-critical learning rates ($\lambda>10^{-3}$), ELRs do not seem to converge anymore, presumably to auto-rate tuning effects becoming too weak compared to fluctuations in base gradient magnitude \cali{caused by training}.

\subsection{ELR Spread and Trainability}
\begin{figure}[tbp]
	\centering
	\includegraphics[width=\linewidth]{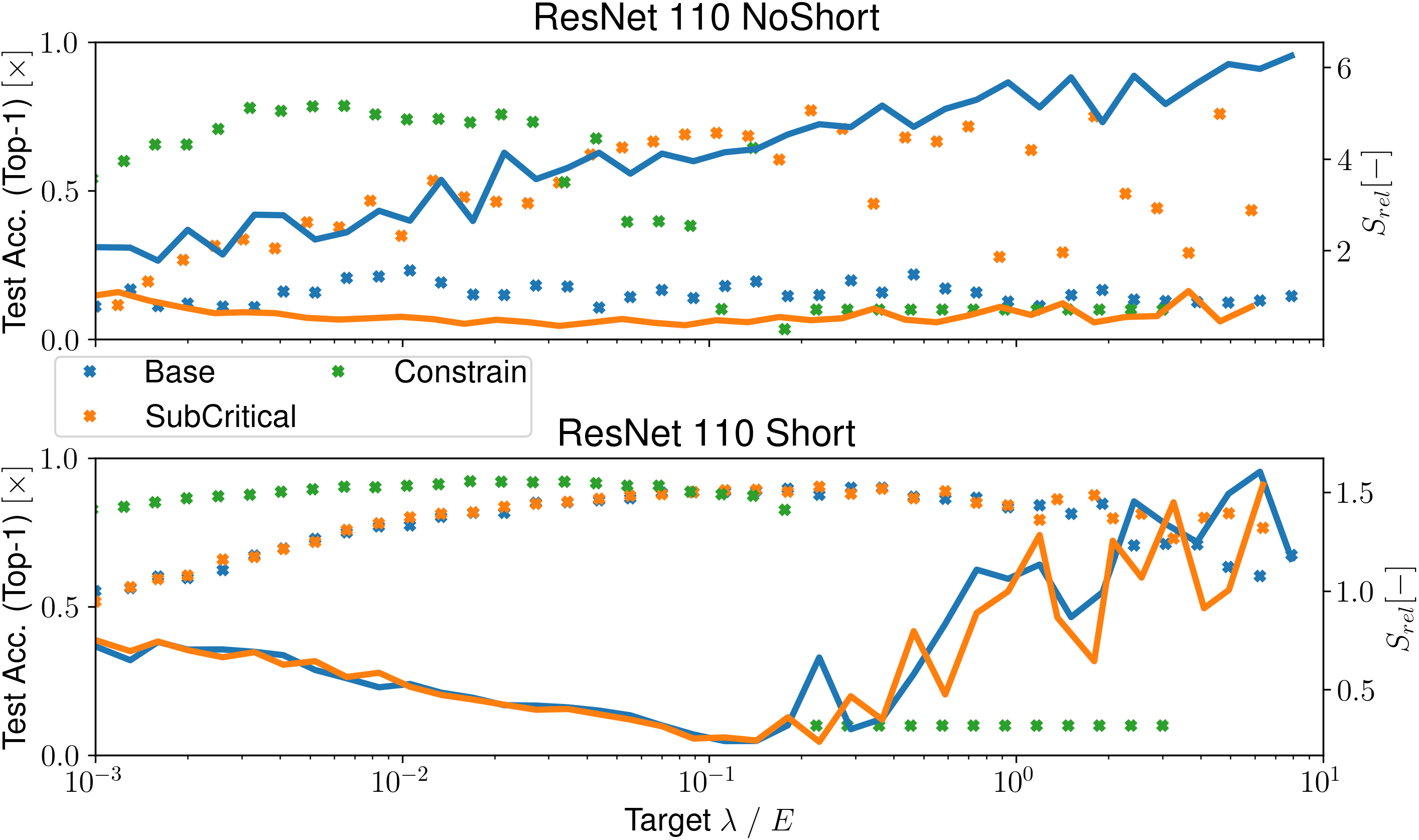}
	\caption{Test accuracies and relative spreads of a ResNet110 (No)Short trained on Cifar10 using regular, warm-up and constrained ELR training protocols for different target (E)LRs.}
	\label{fig:rangetest_all}
\end{figure}
\label{elr_constrain}
In this section, we want to show empirically that networks with high ELR spreads \cali{correlate with} low trainability and that lowering spreads using various methods can restore trainability. For this, we chose an experimental setting where ELR spreads are large: we train a ResNet110 (No)Short on Cifar10. For all runs, we use a simple multistep learning rate decay. In Figure \ref{fig:rangetest_all} (top), we see that for the ``NoShort" networks in regular training without warm-up (``base"), spreads (averaged over the training run) are very high and trainability is very low. Using skip connections (bottom), spreads are much lower and the network is able to train. 

\subsubsection{Subcritical Warm-Up}
As we have seen in Figure \ref{fig:critical_lr} (right), because of learning effects present in real training, the more conservative choice of using the subcritical learning rate for warm-up seems like a more sensible value to avoid overshooting in practice but still guarantees fast convergence in theory  (ref. Corollary \ref{cor:supercritical}). Further, since we are using BatchNorm, we obtain channel-wise ELR values and use the maximum of these values as our layer-wise value.

\subsubsection{Constraining Layer-Wise ELRs}
 Another possibility of controlling ELR spread is scaling each layer's gradients before each step so that layer-wise effective learning rates are constrained to be constant:
\begin{equation}
	\nabla W \leftarrow \nabla W \cdot \frac{E_{goal}}{E+\epsilon},
\end{equation}
for a given constant goal effective learning rate $E_{goal}$ and a small $\epsilon$ we chose as $\epsilon=10^{-5}$. A similar mechanic was used in the popular LARS optimizer \citep{lars}. To prevent increasing weight norms $W$ from overflowing,  we additionally divide all layer weights by the maximum layer weight $\widehat W = max(||W||_F)$ over all layers before every step. This should not change the network function since normalization layers are scale invariant and gradients are normalized. Alternatively, one could re-scale the gradients by $\frac{1}{\sqrt{1+\lambda^2}}$ after every optimization step, as described by Bernstein et al. \yrcite{fromage}. 

\subsubsection{Evaluation}

In Figure \ref{fig:rangetest_all} (left), we see that both techniques lower ELR spread across layers \cali{which correlates with the previously untrainable} ResNet110 NoShort \cali{becoming trainable}, despite its initially exponentially exploding gradients. \cali{Although not a proof of a general causal connection between ELR spread and trainability, the fact that an untrainable network becomes trainable with the intervention made yields some compelling evidence supporting such a hypothesis.} For the network with skip-connections (right), we can observe the same effects but less pronounced, which is expected since the initial spread is linear and not exponential in the number of layers as shown in Section \ref{expl_grads}. In Appendix Section \ref{appendix:sec:additional_experiments}, we repeat this experiment on the Cifar100 dataset and draw similar conclusions. \\

Finally, we train a ResNet101 (No)Short on ImageNet for 50 epochs using three different warm-up schedulers: OneCycle \citep{onecycle}, sub-critical warm-up and no warm-up; we use the exact same cool-down phase (cosine) for all schedulers. We use default hyper-parameters for the OneCycle scheduler that work well in training a ResNet101 Short in a short amount of epochs on this dataset. In Table \ref{fig:imgnet}, we see that indeed warm-up lowers $S_{rel}$ \cali{and correlates with increased} performance, but the hyper-parameters used for OneCycle that work well with the residual network still result in significant spreads for the non-residual network with higher initial spreads. This confirms that warm-up should be scheduled as a function of current ELRs. Although subcritical warm-up uses very few warm-up steps, it yields comparable or better results than our preset OneCycle warm-up. Using the ELR-constrain method to prescribe a global ELR similar to the OneCycle run, we see that we are able to train the network without residual connections; using warm-up additionally decreases performance, showing that warm-up presents no benefits in a setting with no ELR spread.

\begin{table}[h]
\caption{Test accuracies and relative spread of a ResNet101 (No)Short trained on ImageNet using different types of warm-up / ELR-constrain; \textit{RES} indicates residual connections and \textit{CTN} whether the ELR-constrain method was used.}
\label{fig:imgnet}
	\vskip 0.15in
	\begin{center}
		\begin{small}
			\begin{sc}
			\begin{tabular}{l l l l l l}
				\toprule
				Res&Ctn&W. Type&W. Steps&Acc.&$S_{rel}$\\
				\midrule
				No&No&None&0&08.50&3.96\\
				No&No&OneCycle&64060 &22.82&1.96\\
				No&No&Subcritical&9&41.83&0.70\\
				No&Yes&None&0&47.99&-\\
				No&Yes&OneCycle&64060 &45.61&-\\
				Yes&No&None&0&72.85&0.29\\
				Yes&No&OneCycle&64060&72.83&0.31\\
				Yes&No&Subcritical&3&73.06&0.27\\
				\bottomrule
			\end{tabular}
			\end{sc}
		\end{small}
	\end{center}
	\vskip -0.1in
\end{table}

\section{Discussion and Future Work}
In past work, high spreads in effective learning rates have been conjectured to negatively affect trainability, but to our best knowledge, no formal model exists that describes their time-based evolution in early training phases for scheduled learning rates. Under the assumption of constant gradient magnitudes beyond normalization effects, we derived a simple model from first principles that describes the evolution of expected weight/gradient norms and consequently effective learning rates during training. Under our model's assumption, we were able to prove that when training long enough using \textit{any} constant learning rate, all ratios of layer-wise effective learning rates eventually converge to the same value. Problems can still arise in the first step(s) if the learning rate $\lambda(t_i)$ is bigger than the critical value  $\kappa_{\ell h}(t_i)$ (which depends on current weight/gradient norms), momentarily increasing the disparity between layer-wise effective learning rates. We consider this theoretical model of normalization-induced dynamic effects to be our main contribution.\\

In a series of empirical experiments, we have shown that although we exclusively model norm-induced dynamics (scale-invariant linear layers) and assume that the expected gradient norm of other layers (objective function, nonlinear layers) does not change over time, our main takeaway still holds when training a deep convolutional ReLU network on real data: training reduces effective learning rate spread up to a certain \textit{critical learning rate}. By using live gradient values at each step and using a slightly more conservative learning rate choice than predicted, we were able to design a hyper-parameter-free warm-up scheduler that is able to quickly reduce effective learning rate spreads in practice. In an (extreme) setting with exponentially exploding initial gradients, we show \cali{that reducing ELR spreads using warm-up or by normalizing gradients to prescribe a constant effective learning rate correlates with the network's trainability being restored}.\\

\cali{Our analysis applies to all \textit{normalized} networks, i.e. architectures where the network function is invariant wrt. scaling in weight matrices, which is usually the case in most normalized feedforward architectures. Unfortunately, unlike most other traditional MLPs/CNNs/ResNets, the weight matrices of attention blocks are not scale-invariant and thus the inverse scaling property (Eq. \ref{inv_scaling}) and orthogonality (Eq. \ref{orthogonality}), which our model relies on, are violated. Moreover, modifying the architecture (i.e. adding additional normalization layers) would fundamentally impact its way of working (e.g. attention cannot be unlearnt anymore).} Preliminary results show that for architectures containing higher degree nonlinearities (e.g. Transformer models), base gradients can vary much more compared to simple feedforward ReLU networks, therefore limiting the applicability of our model as is. If the order of the fluctuations of the base gradient exceeds that of the auto-rate tuning effects, the effect vanishes. We could envision extending our model to include an error analysis for non-constant base gradients, estimating when this is the case.

\newpage

\section*{Acknowledgments}
The authors acknowledge funding from the Emergent AI Center funded by the Carl-Zeiss-Stiftung. The authors would like to thank Daniel Franzen and Jan Disselhoff for their helpful discussions. The authors would also like to express their gratitude to the HPC working group of the Johannes-Gutenberg University Mainz for sharing their compute power in times of need.

\section*{Impact Statement}
This paper presents work whose goal is to advance the field of Machine Learning. There are many potential societal consequences of our work, none which we feel must be specifically highlighted here.
\Urlmuskip=0mu plus 1mu\relax
\raggedright
\bibliographystyle{plainnat}
\bibliography{bibliography}

\clearpage
\appendix
\section{Additional Proofs}
\label{appendix:sec:additional_proofs}
This section contains proofs of theorems as well as additional material complementary to the main section.

\begin{proof}[Proof of Proposition \ref{prop:independent_layers}]
	We compute the derivative of the output with regard to the input using the chain rule and relate it to the derivative with unscaled inputs:
	\begin{align}
		\frac{dN}{dx} (x,\scalefact W) &= \frac{dN}{dL} (x,\scalefact W) \cdot \frac{dL}{dx} (x, \scalefact W)\\
		&= \frac{1}{\scalefact} \frac{dN}{dL} (x,W) \cdot  \scalefact \frac{dL}{dW} (x, W)\label{eq:3}\\
		&= \frac{dN}{d x} (x, W).
	\end{align}
	Where in Eq. \ref{eq:3}, we used the inverse scaling property from Eq. \ref{inv_scaling}.
\end{proof}	

\begin{proof}[Proof of Theorem \ref{theorem:convergence_bigger_lr}]
	We would like to credit the original author of this proof \cite{mse_proof}. By setting $x_i\coloneq\frac{\sigma^2_j(t_i)}{c_j\lambda}$ and $y_i\coloneq\frac{\sigma^2_k(t_i)}{c_k\lambda}$, we can rewrite Eq. 4 for layers $j$ and $k$ as two sequences obeying the same recurrence relation:
	\begin{align}
		x_{i+1} &= x_i + \frac{1}{x_i}\\
		y_{i+1}& =  y_i + \frac{1}{y_i}.
	\end{align}
	Raising $x_i$ to the second power yields:
	\begin{equation}
		x_{i+1} ^2 = x_i^2 + 2 + \frac{1}{x_i^2}.
	\end{equation}
	This allows us to unroll the recursion as follows : 
	\begin{equation}
		x_i^2 = x_1^2 + 2(i-1) + \frac{1}{u_1^2} + \ldots + \frac{1}{u_{i-1}^2}.
	\end{equation}
	As $x_j \geq 2(j-1)$, we can write the following inequality:
	\begin{equation}
		2(i-1) \leq x_i^2 \leq 2(i-1) + x_1^2 + \frac{1}{x_1^2} + \frac{1}{2} + \frac{1}{4} + \ldots + 2(i-2).
	\end{equation}
	By the integral test, it is clear that $\sum_{i=1}^{n-1} \frac{1}{k} \leq ln(n)$ and therefore $\sum_{i=1}^{n-1} \frac{1}{2k} \leq  \frac{ln(n)}{2} = ln(\sqrt{n})$. Let be \text{$\gamma\coloneq u_1^2+\frac{1}{u_1^2}-2$}, we consider the square root of the expression above:
	\begin{equation}
		\sqrt{2i-2}\leq x_i \leq \sqrt{2i+ log(\sqrt{i-1})+\gamma }.
	\end{equation}	 	
	Since $\gamma$ is a constant and  $\lim_{i\rightarrow \infty}\frac{log(i)}{i} = 0$, it follows that:
	\begin{equation}
		\lim_{i\rightarrow\infty} \frac{x_i}{\sqrt{2i}} = 1.
	\end{equation}
	and analogously
	\begin{equation}
		\lim_{i\rightarrow\infty} \frac{y_i}{\sqrt{2i}} = 1.
	\end{equation}
	We therefore obtain:
	\begin{equation}
		\lim_{i\rightarrow\infty} \frac{x_i}{y_i}  = \frac{\sigma^2_j c_k}{\sigma_k^2c_j}(t_i)= R_{k j} (t_i) =1,
	\end{equation}
	which is in turn also true for the inverse fraction.
\end{proof}

\begin{proposition}[Ratios Flip at Most Once]
	\label{prop:fliponce}
	Let $j, k \leq L $ be any layer pair with w.l.o.g. $R_{j k}(t_i)> 1$ and assume a constant learning rate $\lambda(t_i) = \lambda$.
	\begin{enumerate}
		\item 	If effective learning rate ratios do not flip between a given time step and the next, they will never flip at a later time step, i.e.  if $R_{j k} (t_{i+1}) > 1$ it follows that  $R_{j k} (t_{i+j}) > 1$ for all $j \geq 1$.
		\item 	If effective learning rate ratios do flip between a given time step and the next, they will never flip again at a later time step,, i.e.  if $R_{j k} (t_{i+1}) < 1$ it follows that  $R_{j k} (t_{i+j}) < 1$ for all $j \geq 1$.
	\end{enumerate}
	
\end{proposition}

\begin{proof}
	We start by showing the first statement. Assuming that the effective learning rate ratio does not flip between time steps $t_i $ and $t_{i+1}$, we know by Lemma \ref{lemma:flipping_conditions} that $\lambda < \kappa_{j k}(t_i)$. We now just have to show that $\lambda < \kappa_{j k}$ for all successive time steps. Since $c_j$ and $~c_k$ are constants and we know by the definition of the discrete process in Eq. \ref{discrete_model_weights} that all weight norms $\sigma(t_i)$ are strictly increasing over time, we can write:
	\begin{equation}
		\lambda < \kappa_{j k}(t_{i}) = \frac{\sigma_j\sigma_k}{\sqrt{c_j c_k}}(t_{i}) < \frac{\sigma_j\sigma_k}{\sqrt{c_j c_k}}(t_{i+j}) = \kappa_{j k}(t_{i+j})
	\end{equation}
	for all $j\geq 1$ and thus by Lemma \ref{lemma:flipping_conditions} the ratio will never flip again.\\
	We now show the second statement.  Assuming that the effective learning rate ratio does flip between time steps $t_i $ and $t_{i+1}$, we know by Lemma \ref{lemma:flipping_conditions} that $\lambda > \kappa_{j k}(t_i)$. We start by showing that the ratio will not flip for the next time step, which is in turn equivalent to $\lambda < \kappa_{j k}(t_{i+1})$. We can expand this term as follows:
	\begin{align}
		\kappa_{j k}^2(t_{i+1}) &= \frac{\sigma^2_j\sigma^2_k}{c_j c_k}(t_{i+1})\\
		&= \frac{1}{c_j c_k}\left(\sigma_j^2+\frac{c_j^2\lambda^2}{\sigma_j^2}\right) \left(\sigma_k^2+\frac{c_k^2\lambda^2}{\sigma_k^2}\right)(t_i)\\
		&=\left( \frac{\sigma_j^2\sigma_k^2 }{c_j c_k}+ \frac{\sigma_j^2 c_k˝\lambda^2}{\sigma_k^2c_j} + \frac{\sigma_k^2 c_j\lambda^2}{\sigma_j^2 c_k} + \frac{c_j c_k\lambda^4}{\sigma_j^2\sigma_k^2}\right)(t_i)\\
		&= \left(\kappa_{j k}^2 + \frac{E_k}{E_j} \lambda^2+ \frac{E_j}{E_k} \lambda^2 + \frac{1}{ \kappa_{j k}^2} \lambda^4\right) (t_i) \\
		&>\left(\kappa_{j k}^2 + \frac{E_k}{E_j} \lambda^2+ \frac{E_j}{E_k} \lambda^2 + \frac{1}{ \lambda^2} \lambda^4\right) (t_i)\label{eq:1}\\
		&\geq \lambda^2 \label{eq:2}.
	\end{align}
	We can write Eq. \ref{eq:1} because of the assumption that $\lambda > \kappa_{j k}(t_i)$ and Eq. \ref{eq:2} because all summands are non-negative. We have therefore shown that $\lambda < \kappa_{j k}(t_{i+1})$ and thus the ratio will not flip between time step $t_{i+1}$ and $t_{i+2}$. By the first proposition shown above, we know it will therefore never flip in future time steps, i.e. $R_{j k} (t_{i+j}) < 1$ for all $j \geq 1$.
\end{proof}

\section{Additional Experiments}
\label{appendix:sec:additional_experiments}
\begin{figure}[tbp]
	\centering
	\includegraphics[width=\linewidth]{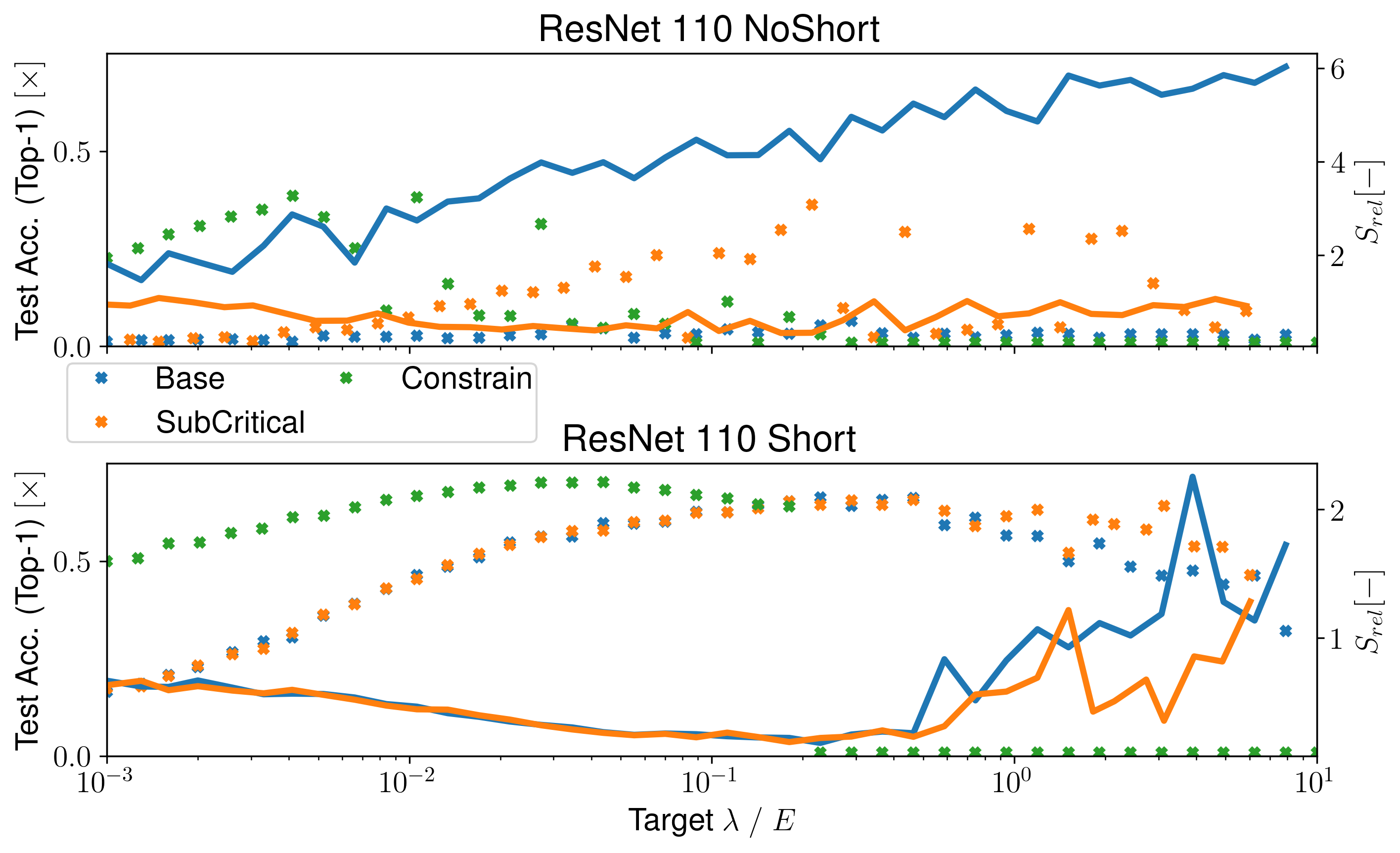}
	\caption{Test accuracies and relative spread of a ResNet110 (No)Short trained on Cifar100 using regular, warm-up and constrained ELR training protocols.}
	\label{appendix:fig:rangetest_all_cifar100}
\end{figure}

In Figure \ref{appendix:fig:rangetest_all_cifar100}, we repeated the experiment of Figure \ref{fig:rangetest_all} on the Cifar100 dataset. Qualitatively, we observe the same effects. The ResNet110 NoShort does not train at all and has high ELR spreads. By using the ELR-constrain or critical-warmup method, we are able to train the network to a significant, but not very good performance. As for the ResNet110 Short, we start to see a difference between runs without warm-up our sub-critical scheduler for high learning rates $\lambda>10$) where again, a reduced spread results in increased trainability. We conclude that reducing ELR spread \cali{correlates with increased trainability}, but other factors (e.g. vanishing dimensionalty) explain the gap between short and no-short architectures.

\section{Other Factors Influencing Weight Dynamics}
\label{appendix:sec:limitations}
As mentioned in the main paper, some techniques commonly used in training influence the evolution of weight dynamics \cali{in a way that is not modeled by Eq. \ref{discrete_model_weights}}; in this section we will discuss them.

\subsection{Weight Decay}
The fact that weight decay influences weight dynamics in normalized networks is quite trivial and well-explored in recent literature: \cite{norm_matters} \cite{laarhoven}. In a normalized network, if all weights are reduced by a factor $\alpha$, this corresponds to an increase of the global learning rate by a factor $\alpha$, as per Eq \ref{inv_scaling}.

\subsection{Momentum}
As momentum SGD \cite{momentum_sgd} modifies each gradient's direction and length before it is applied, it is easy to see that it must influence weight dynamics. It is possible to compute weight dynamics of a network optimized with momentum SGD, but we consider this to be out of scope of this work.

\subsection{Affine Normalization Parameters}
\begin{figure}[tbp]
	\centering
	\includegraphics[width=\linewidth]{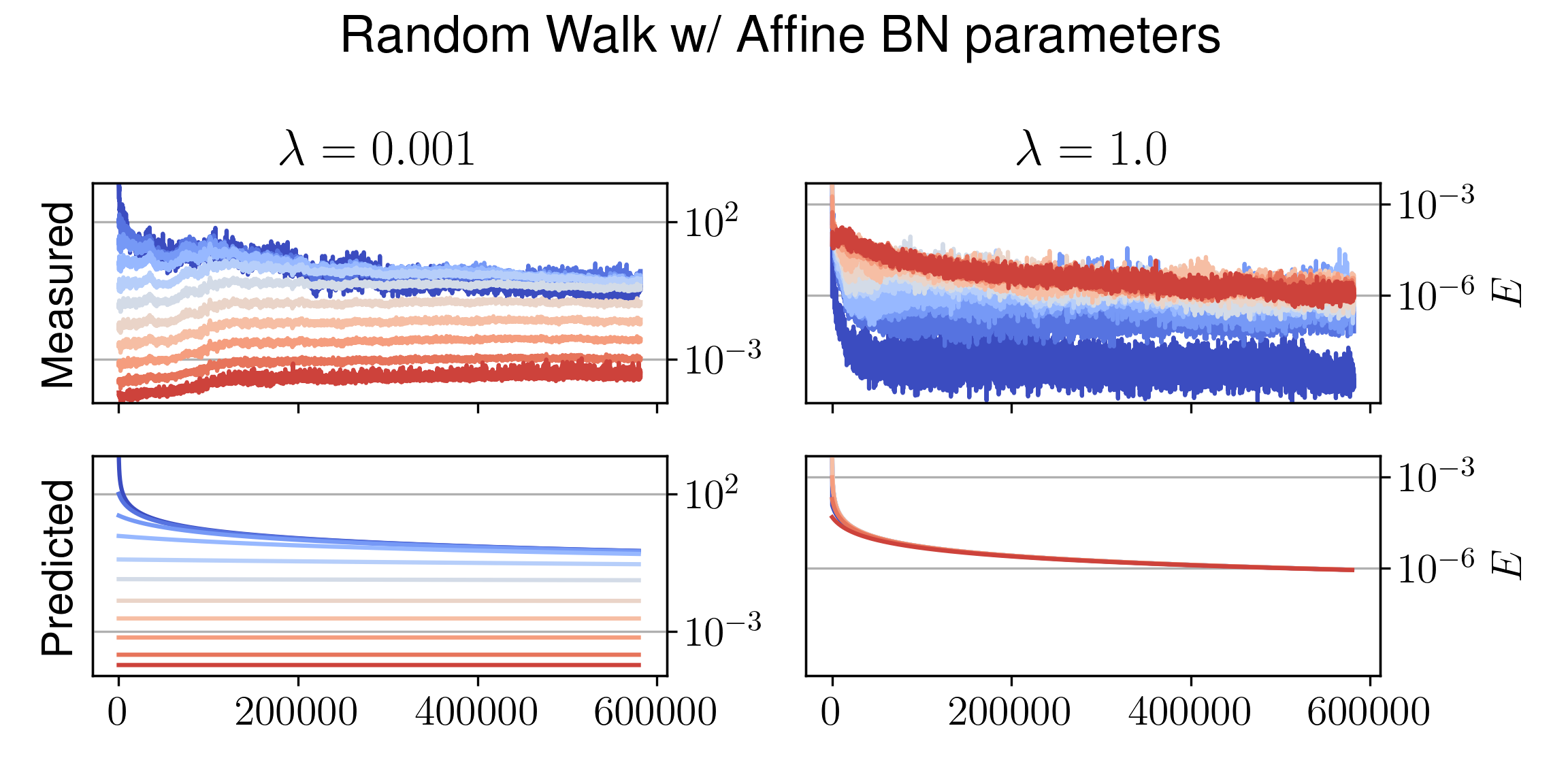}
	\caption{Long term evolution of simulated/real layer-wise effective learning rates for a ResNet56 NoShort trained with random gradients and affine BatchNorm parameters.}
	\label{appendix:fig:affine}
\end{figure}

Normalization layers are usually applied with learnable affine parameters \text{$\gamma\cdot N(x)+\beta$} that are initialized to $\gamma=1$ and $\beta=0$ \citep{torch_batchnorm2d}. In the case of a network where normalization layers are followed by ReLUs (this is the case in our experiments), this means that we initialize in the ``maximum curvature region" of the nonlinearity but drift away from it during training \citep{ringing_relus} leading to gradients dropping further than expected.  In Figure \ref{appendix:fig:affine}, we repeated the experiment of Figure \ref{fig:pred_drift} using random gradients (a setting that produces a reliable prediction) but add affine BatchNorm parameters to our training protocol. For $\lambda=0.001$, the prediction is still quite accurate but for $\lambda=1$, we see that the real gradients are much smaller than predicted.

\section{Architecture and Training Details}
\label{sec:arch_details}
As described in the main paper, we used ResNet variants with varying hyper-parameters, with and without skip connections. Architectural details can be found in the tables below.

The experiments in the paper were made on computers running Arch Linux, Python 3.11.5, PyTorch Version 2.1.2+cu121. Various Nvidia GPUS were used ranging from GeForce GTX 1080TI, GeForce GTX 2080Ti RTX 4090.

\begin{table}[tbp]
	\caption{Network architecture and training regime used for the Cifar10/100 task.}
	\begin{center}
		\begin{small}
			\begin{sc}
					\begin{tabular}{ l  l }
				\toprule\\
				Architecture& ResNet56/110\\
				\midrule\\
				Block & BasicBlock v1\\
				Num. Blocks & 9 9 9 / 18 18 18\\
				Num. Planes & 16 32 64\\
				Shortcut Type & A (Padding)\\
				\bottomrule
			\end{tabular}
			\end{sc}
		\end{small}
	\end{center}
	\hspace{3in}
		\begin{center}
		\begin{small}
			\begin{sc}
				\begin{tabular}{ l  l }
					\toprule\\
					Training& CIFAR-10 / CIFAR-100\\
					\midrule\\
					Epochs& 200\\
					Scheduler&Multistep ($\gamma=0.1$)\\
					Milestones&100, 150\\
					Learning rate&Variable\\
					Batch size&256\\
					Optimizer& SGD\\
					Momentum& 0\\
					Weight decay &0\\
					Augmentation& Random Flip\\
					Nesterov& False\\
					\bottomrule
				\end{tabular}
			\end{sc}
		\end{small}
	\end{center}
\end{table}

\begin{table}[htbp]
	\caption{Network architecture and training regime used for the ImageNet task.}
		\begin{center}
		\begin{small}
			\begin{sc}
				\begin{tabular}{ l  l }
					\toprule\\
					Architecture& ResNet101\\
					\midrule\\
					Block & BottleneckBlock v1\\
					Num. Blocks & 3 4 32 3\\
					Num. Planes & 64 128 256 512\\
					Shortcut Type & B (1x1-Conv+BN)\\
					\bottomrule
				\end{tabular}
			\end{sc}
		\end{small}
	\end{center}
		\hspace{3in}
	\begin{center}
		\begin{small}
			\begin{sc}
	\begin{tabular}{ l  l }
	\toprule\\
	Training& ImageNet\\
	\midrule\\
	Epochs& 50\\
	Scheduler&OneCycle/\\
	&No-Warmup + Cosine/\\
	&Subcritical + Cosine\\
	Max. LR&0.4\\
	Batch size&100\\
	Optimizer& SGD\\
	Nesterov& False\\
	Momentum&0\\
	Weight decay& 0\\
	Augmentation& Random Flip\\
	\midrule
	OneCycle AnnealStrategy& Cosine\\
	OneCycle BaseMomentum& 0\\
	OneCycle CycleMomentum& True\\
	OneCycle DivFactor& 20\\
	OneCycle EpochsStart& 0.1\\
	OneCycle FinalDivFactor& 2000\\
	OneCycle MaxMomentum& 0.0 \\
	\bottomrule
\end{tabular}
			\end{sc}
		\end{small}
	\end{center}
\end{table}

\end{document}